\documentclass[journal]{IEEEtran}
\usepackage{amssymb}
\usepackage{mathptmx}                            
\usepackage{amsmath}
\usepackage{amsfonts}
\usepackage{color}
\usepackage{subfig}
\usepackage{graphicx}
\usepackage{siunitx}
\usepackage{color}
\usepackage{textcomp}
\usepackage{lscape}
\usepackage{ctable}
\usepackage{algorithmic}
\usepackage{algorithm}
\usepackage{tabularx}
\usepackage{multirow}
\usepackage{makecell}
\usepackage[square,numbers]{natbib}
\usepackage{float}
\graphicspath{{./Figures/}}                             
\usepackage[english]{babel}
\usepackage[utf8]{inputenc}
\usepackage{fancyhdr}
\usepackage{multicol}

\ifCLASSINFOpdf
\else
\fi

\begin{document}

\twocolumn[  
    \begin{@twocolumnfalse}

            This work has been submitted to the IEEE for possible publication. Copyright may be transferred without notice, after which this version may no longer be accessible.

     \end{@twocolumnfalse}
]
\title{Adaptive Surgical Robotic Training Using Real-Time Stylistic Behavior Feedback Through Haptic Cues}

\author{Marzieh~Ershad, Robert Rege, and Ann~Majewicz~Fey
\thanks{M. Ershad, the Department of Electrical Engineering, University of Texas at Dallas, Richardson, TX, 75080. e-mail: marzieh.ershadlangroodi@utdallas.edu}
\thanks{A. Majewicz Fey, the Department of Mechanical Engineering, University of Texas at Austin, Austin, TX 78712}
\thanks{A. Majewicz Fey and R. Rege, the Department of Surgery at UT Southwestern Medical Center, Dallas, TX, 75390}
}

\maketitle

\begin{abstract}
Surgical skill directly affects surgical procedure outcomes; thus, effective training is needed to ensure satisfactory results. Many objective assessment metrics have been developed and some are widely used in surgical training simulators.
These objective metrics provide the trainee with descriptive feedback about their performance however, often lack feedback on how to proceed to improve performance. The most effective training method is one that is intuitive, easy to understand, personalized to the user and provided in a timely manner.
We propose a framework to enable user-adaptive training using near-real-time detection of performance, based on intuitive styles of surgical movements (e.g., fluidity, smoothness, crispness, etc.), and propose a haptic feedback framework to assist with correcting styles of movement.
We evaluate the ability of three types of force feedback (spring, damping, and spring plus damping feedback), computed based on prior user positions, to improve different stylistic behaviors of the user during kinematically constrained reaching movement tasks.  
The results indicate that four out of the six styles studied here were statistically significantly improved (p$<$0.05) using spring guidance force feedback and a significant reduction in task time was also found using spring feedback.
Path straightness and targeting error were other task performance metrics studied which were improved significantly using the spring-damping feedback.
This study presents a groundwork for adaptive training in robotic surgery based on near-real-time human-centric models of surgical behavior. 
\end{abstract}


\IEEEpeerreviewmaketitle

\section{Introduction}
Surgical outcomes are highly dependent on surgeon skill levels. Efficient training that provides trainees with appropriate feedback and assists them with achieving expert-like performance is critical for mastering technical skills in surgery and achieving successful procedural outcomes~\cite{curry2012}. 
Traditional methods in surgical training typically involve an expert observing and evaluating a trainee's performance in the operating room, and providing feedback to the trainee on how to improve performance~\cite{cameron1997}. 
Automating skill assessment can alleviate the time intensiveness and subjectiveness of these methods; Furthermore, finding an effective and efficient feedback method, which is intuitive and easy to understand is crucial~\cite{hoffman2015}.

For patient-free and more objective training environments, virtual reality (VR) simulators have begun to find their way into surgical training~\cite{gallagher2005, badash2016}. Simulators provide factual and quantitative data to the human user upon completion of each simulated task, such as number of instrument collisions, time to complete the task, and the number of missed targets. These metrics indicate the success rate of the trainee but do not necessarily provide them with meaningful feedback on how to modify their movements to improve performance~\cite{sewell2008}. 

To address this issue, an ongoing development in surgical simulators is to enable real-time feedback to users based on the calculated metrics. Errors are computed from the user's interaction with the virtual environment with respect to a reference, and to correct this error,  feedback is provided to the user accordingly.

The error can be calculated based on deviating from an expected trajectory or performance variable.
Haptic feedback has been widely used for training purposes in simulators to assist with following a specified trajectory or providing a sense of touch in interaction between the tissue and the instrument while performing a task. An example of haptic feedabck for providing tissue-tool interaction is the work from  Pezzementi et al. They implemented a platform for interaction with soft tissue in a simulated environment using the Phantom Omni haptic device by training a linear 2D mass-spring-damper system which performs similar to a nonlinear finite element (FE) model~\cite{pezzementi2008}. For trajectory guidance, Ko et al. developed a training simulator to assist the trainee with following a desired catheter insertion path through haptic feedback by calculating forces during catheter insertion~\cite{ko2017}. These methods have proved to be effective in improving performance however, do not incorporate user's behavior or movement style which provide rich information regarding user's proficiency, and can lead to more intuitive training methods.

An effective training method should be easily interpretable by the user. In an earlier study~\cite{Ershad2016}, we showed that the quality of movement during task performance which is intuitively perceived by a human observer can be used to distinguish different expertise levels;
thus the user's style of movement includes valuable information regarding his/her skill level, and deviations from expert-like movements can be used to calculate relevant feedback for training.


Another recently explored source for improvement in virtual reality surgical simulators is adaptive training which provides relevant and customized training feedback to trainees, based on individual strengths and weaknesses, and could enhance learning outcomes.
The large amount of data recorded and stored by VR simulators enables data-driven analysis and automatic performance evaluation. This enables adaptive training based on each individual's performance~\cite{vaughan2016}. An example of an adaptive robotic surgical training framework is presented in~\cite{mariani2018}. This study compares adaptive curriculum training to self-managed training and shows significant improvement in performance and learning skill using an adaptive framework. However, these performance assessment and adaptive feedback methods are largely task-dependent, which limit the generalizability of these approaches. 

In the following, we will discuss previous studies in this field and describe our proposed methodology which addresses the issues mentioned above to assist with improving training in robotic surgery.
The rest of the paper is structured as follows. In section~\ref{relatedWork} we summarize related work in adaptive training, force-reflective feedback, and guidance force feedback. In section~\ref{methods} our proposed stylistic assessment and feedback method is discussed in detail. It includes a deficiency detection phase and a feedback applying phase. A deficiency in style is detected from user hand position and velocity data, by comparing to expert style for a variety of stylistic descriptors. Then, the user is provided with either spring, damping, or spring-damping force feedback, depending on their randomized assignment of our feedback groups. We evaluate the effectiveness of our adaptive stylistic force feedback using both performance metrics as well as stylistic changes over the duration of the experimental study. Section~\ref{human study} describes the experiment design and tools used to conduct the experiment. In section~\ref{results}, we present the results of the proposed training method, and discuss the effect of the different types force feedback on styles of movement. Section ~\ref{conclusion} concludes the paper and suggests the future research in this field.

\section{Related Work}
\label{relatedWork}
\subsection{Adaptive Training}
Adaptive technology can be introduced into training devices to develop user-specific training that results in more effective learning. 
An adaptive system can be known as a supervisor that instructs each trainee based on his/her unique performance and provides specific instructions on how to proceed, or adjusts the training task for each individual to ensure best results for each trainee. These systems consist of a control loop that detect changes in the output from a desired point, this can be done using machine learning approaches that enable deficiency detection or performance classification . Feedback is then applied to modify the response to move it towards the desired performance levels~\cite{vaughan2016}. Adaptive systems require three main elements including constant monitoring and measurement of performance, an adaptive variable, and a methodology to adjust the variable to enhance performance~\cite{kelley1969}.

In adaptive training, the user's performance is evaluated based on specific criteria (detection phase) and then in the next step training is adapted accordingly (feedback or training phase). Task difficulty level is one element of focus in the training phase. The difficulty of the task can be updated based on user's performance to adjust the level of challenge and enhance learning. This has been studied in digital games~\cite{charles2005}. The stimulus or the type of feedback provided to the user is another element of focus in the training phase in which the feedback is adapted based on the user's performance. Visual, audio, and haptic feedback are some of types of feedback used in the training phase.
Different types of haptic feedback used in training systems, will be discussed in the following section.

\subsection{Haptic Feedback for Training}
To study the effect of haptic feedback on user's performance, two types of haptic feedback are noticeable: reflective feedback and guidance feedback. The former provides the user with a feeling of touch and force in interacting with an object in environments where these senses are missing. In virtual environments, this is done through haptic rendering.  The latter provides the user with haptic cues, and assists in correcting movements to improve performance.

\subsubsection{Reflective Feedback}
Haptic feedback in VR simulators improves training~\cite{basdogan2004}.
The lack of haptic feedback (both force and tactile) causes an inappropriate level of force applied to the tissue which can lead to safety issues~\cite{enayati2016}. Tactile feedack decreases the force applied to the tissue and hence reduces tissue damage. A study was conducted to show this effect on robotic manipulation by mounting force feedback system onto a da Vinci surgical robotic system performing multiple peg transfer tasks~\cite{king2009}. This study showed that higher force was applied by all subjects in the absence of haptic feedback; thus, indicating that tactile feedback assists surgeons with tissue handling by applying an appropriate amount force to the tissue. In another study, the effect of tactile force feedback was evaluated in vivo~\cite{wottawa2016}. This study also showed a significant reduction in grasping forces and thus, tissue damage in the presence of an integrated tactile feedback.
A study by Abiri et al. showed that a multi-modal feedback including tactile, kinesthetic, and vibrotactile feedback for providing a sense of touch in tissue grasping and manipulation tasks resulted in an average of 50\% reduction in force compared to a no feedback scenario~\cite{abiri2019}.

This type of reflective feedback though proving to be helpful in providing the user with a feeling of touch and force in teleoperated environments where these sense are missing, do not provide any cues to the user on how to modify movement to improve performance. Guidance feedback which addresses this issue is discussed in the following.

\subsubsection{Guidance Feedback}
Haptic guidance can enhance learning new motor skills in robotic environments where an instructor is not present to guide the user on how to modify his/her movement. Different studies have shown the effectiveness of haptic feedback in developing motor skills~\cite{Boulanger2006}, as well as movement guidance~\cite{6226397}.
A common type of training motor skills using haptic feedback, is transferring expert skills in which a an expert's movements are recorded and played back to train a novice~\cite{Yang2008}. 
However, Gibo et al. showed that haptic feedback can help discover new movement strategy rather than following a specific trajectory or enforcing a specific movement~\cite{gibo2016}. They provided the subjects with an environment to explore different types of movement using haptic feedback and adopt the best strategy.
Haptic disturbances which suggests disturbing the movemnet instead of guding the user can also improve motor skills~\cite{lee2010}.

while all these methods prove the effectiveness of haptic feedback in movement guidance, they do not focus on performance feedback. Jantscher et al. designed and implemented a framework that provides vibrotactile feedback method based on movement smoothness. They proved that the smoothness-based
feedback improved accuracy compared to trajectory based feedback methods~\cite{8357183}. 
They provided the subjects with vibrotactile cues with a degree of pleasantness relative to their performance; however, results in the literature show scenarios in which force feedback assists the user with performing a task, yet is perceived negatively by the human user~\cite{gwilliam2009}, and other scenarios in which force feedback does not improve performance, yet is preferred~\cite{mcmahan2011}.
These results indicate that objective performance metrics and subjective user response surveys may not be sufficient for understanding the intuitiveness of a control interface. 

Similar to~\cite{8357183}, to further investigate performance based haptic feedback, we examine the effect of haptic cues on six different stylistic performance behaviors and study three different types of feedback to find the best type which improves each movement style. 
We propose a framework to provide task-independent stylistic feedback to the human user during movement-based training tasks to provide the user with a more intuitive and global understanding of their movement styles. We designed, implemented, and evaluated an adaptive training method composed of the following elements: (1) Our proposed framework first evaluates the user's stylistic behavior performance in near real time and detects deficiencies in some movement styles~\cite{ershad2018a}. (2) Next, it provides the user with haptic cues to modify their movement to improve performance. We also evaluate the effectiveness of three common types of haptic feedback, namely, spring, damping and spring-damping feedback that is computed from prior user positions and velocities. 
The goal of our study is to find intuitive ways to communicate with the user on how to modify his/her movement to enhance performance. We evaluate the user performance, based on the quality of movement through monitoring their styles of movement (movement styles are is described in section~\ref{methods}.A) while performing a task. We then provide haptic feedback to the user to help correct their style in near-real-time.

\section{Methods}
\label{methods}
Our goal is to improve robot assisted training to help achieve mastery in surgical robotics. For this purpose, we  aim to (1) introduce a customized framework in which each individual is provided training based on his/her performance, (2) provide the trainee with feedback in a timely manner and in near-real-time, (3) introduce a generalizable and task-independent framework which evaluates performance based on the user's style of movement, and (4) develop a more understandable and intuitive way to communicate with the user on how to modify movement to improve performance. 

\begin{figure*}[tb]
\centering
\includegraphics[width=0.8\linewidth,clip ,trim={0mm 0mm 0mm 0mm}]{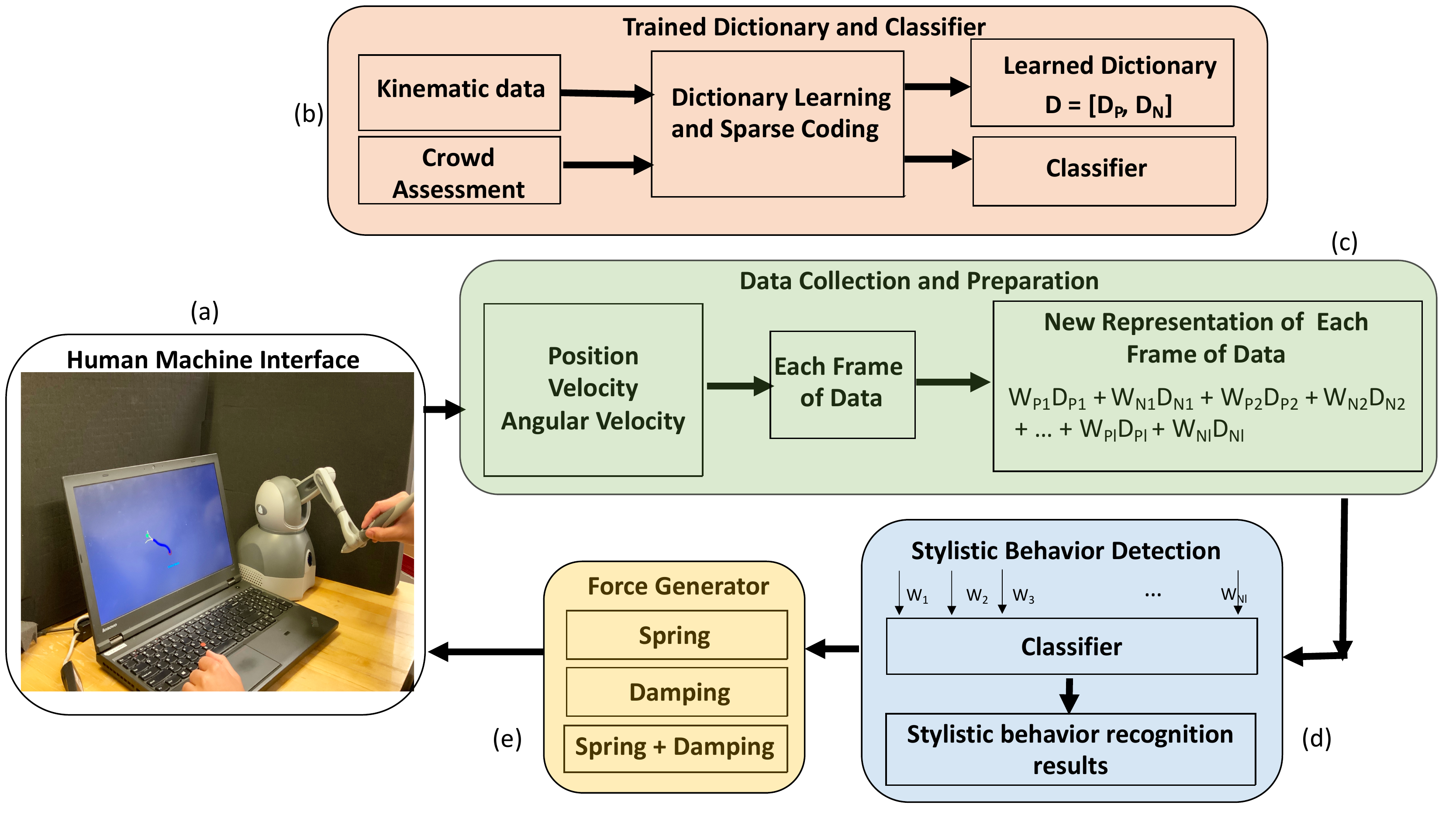} \\
\caption{System Block Diagram: The human user interacts with a haptic device and the simulation environment (a). In advance of the experiment, training movement data is used to learn a dictionary of stylistic features and a classifier is trained to predict stylistic deficiencies in near-real-time~\cite{ershad2019} (b). During the experiment, kinematic measurements from the haptic device is represented into stylistic behaviors by projecting it on the learned dictionary (c). The quality of the user's style is detected using a classifier which takes the coefficients of the new representation of the data as an input (d). Finally, force feedback is provided to the user if negative performance is detected. Three different types of feedback forces were evaluated in this study for their effectiveness in improving user style. Feedback is computed from prior user positions and velocities (e).} 
\label{block-diagram}
\end{figure*}

A systematic framework for recognizing the quality of movement through stylistic behavior and applying appropriate feedback for correcting the style was developed using a human machine interface (i.e., a haptic device) and a simulated task. Fig.~\ref{block-diagram} shows the block diagram of the proposed method. 

\subsection{Surgical Skill Assessment Using Stylistic Behavior}
In a previous study, we presented a novel surgical skill assessment method based on the user's stylistic behavior~\cite{Ershad2016}. These styles of movement represents the appearance of movement in action described by common adjectives used to describe the quality of movement such as smoothness, fluidity, decisiveness, etc., which is easily distinguishable to a casual observer with different expertise levels.
The idea behind this method is that the quality of movement holds fundamental information about a subject's skill; thus, quantifying these universally understandable movement descriptors enables the development of effective and intuitive training strategies.
We proposed a lexicon of contrasting adjectives representing surgical styles through collaboration with expert surgeons (Table~\ref{tab-lexicon}).
To evaluate the ability of these stylistic descriptors in differentiating among different expertise levels, we used crowd-sourced assessment which has proven to show comparable results to expertise evaluation. Paired videos of a subject performing a simulated surgical task and the task being performed was posted to Amazon Mechanical Turk and the crowd rated the videos based on the stylistic descriptors.

To quantify the qualitative assessment based on stylistic behavior, we found data metrics associated with each stylistic behavioral adjective in the lexicon through an extensive search among different calculated metrics. For each adjective, we found the metric that correlated best with the crowd ratings. These metrics were calculated from kinematic and physiological measurements recorded from multiple sensors from the user's hand movement while performing a simulated task on the da Vinci surgical simulator.
Furthermore, we evaluated the ability of the stylistic descriptors to differentiate between different expertise levels. For this purpose, the metrics associated with the stylistic behavior were used to train a classifier which was then used to distinguish among four levels of expertise (novice, intermediate, expert, fellow)~\cite{ershad2018a}. The results showed that these styles of movement were able to distinguish among different expertise levels.

In the next step, to avoid the feature engineering required in the previous study for identifying the stylistic behavior and to detect the deficiency in the styles of movement during user's performance, we proposed an automatic method for extracting underlying structures that represent stylistic behavior from raw kinematic data within 0.25 seconds of movement~\cite{ershad2019}. 

In this study, we design an experiment, to implement, and test the framework for automatically detecting the deficiency in movement styles in near real-time. In addition, to assist with correcting the style of movement as a ground work for developing a training framework, we examine the effect of haptic guidance using three different types of force feedback (spring, damping, spring and damping) on the six different styles of movement in Table~\ref{tab-lexicon}.
\begin{table}[h]
\small
\centering
\caption{Lexicon of Stylistic Behavior}
\begin{tabular}{|c|c|}
	\hline
	\hline
	\multicolumn{1}{|c|}{Positive Adjective} & \multicolumn{1}{|c|}{Negative Adjective} \\
	\hline
	
	{Fluid} & {Viscous} \\	
	{Smooth} 	& {Rough}\\
    {Crisp} 	& {Jittery} \\	
    {Relaxed} 	 & {Tense} \\
    {Deliberate}  & {Wavering}  \\	
	 {Coordinated} 	& {Uncoordinated} \\	
	\hline
	\hline
\end{tabular}
\\
\label{tab-lexicon}
\end{table}

\subsection{Detecting Deficiencies in Stylistic Behavior}
\label{dictionary}
A framework for detecting the stylistic behavior performance is described in~\citep{ershad2019}. A similar approach is used in this study, however, a different data set is used and the model tuned to best fit the new data set. This approach is discussed in the following.

\subsubsection{Crowd-Sourced Assessment for Positive and Negative Performance for Each Style} 
\label{Crowd_jigsaws}
To be able to train a model to recognize a deficiency in movement styles, we first label the data based on a positive or negative performance of the stylistic behavior. For this purpose, we use the JIGSAWS data set~\cite{gao2014jhu}, which is a publicly available data set that contains robotic surgical training videos and kinematic recordings. JIGSAWS videos were uploaded to Amazon Mechanical Turk and crowd workers rated the videos based on the quality of performance in the six styles of movement mentioned in Table~\ref{tab-lexicon}. The crowd-workers were asked to rate the video based on either a positive or negative adjective for a given stylistic descriptor (e.g., smooth v.s. rough movement, crisp vs. jittery movement). Each video was rated by 20 crowd-workers. The trial was eventually assigned a positive label if it was rated positive by more than or equal to 50 \% of the crowd-workers and was otherwise assigned as negative. 

\subsubsection{Dictionary Training and Classifier Model Training} 
\label{Detection}
For each stylistic behavior, a dictionary containing the basis vectors for the good and bad performance is created using the kinematic data from the right hand manipulator of da Vinici skill simulator from the JIGSAWS data set. This data set includes position, velocity, and angular velocity from the robot end effectors.
A separate dictionary was learned from the positive as well as the negative performance and then the total dictionary was obtained from the concatenation of the these two sets of dictionary such that the first half of the basis vectors were dictionary learned from the good performance and the second half were the dictionary learned from the bad performance. 
The positive and negative labels regarding each stylistic behavior adjective used to train the model were obtained from crowd-sourced assessment on the JIGSAWS video data set (section~\ref{Crowd_jigsaws}). The input data is then represented as a linear combination of the basis vectors in the dictionary. The dictionary and the coefficients are calculated using an optimization algorithm that iterates between two problems: 1) finding the basis vectors such that the reconstructed signal is as similar as possible to the input signal, and 2) finding the coefficients such that they are sparse. The sparseness reduces the computational complexity and enables near-real-time implementation. 
These sparse codes are then used to train a SVM classifier. Six separate codebooks are learned for each of the six stylistic behavior adjectives, leading to six trained classifiers.
%

\subsubsection{Coefficient Calculation}
For a new set of input data (i.e. a new frame of 30 samples), dimensionality reduction is done using principle component analysis (PCA) to remove correlations in the data set, then this reduced dimension data set is projected onto the learned codebook (described in section~\ref{Detection}). The new representation of the input signal is sparse. The sparse codes form the new data frame at each point of time, which are then fed into the trained classifier (described in section~\ref{Detection}) for performance evaluation.
Algorithm 1 shows the pseudo code for this method.

\begin{algorithm}
    \caption{Style Performance Detection Algorithm}
   \textbf{Input:} new data  \\
        \textbf{Output:} stylistic behavior performance $S_i$ \\
          \begin{algorithmic}[1]
          \WHILE {trial not finished}
    \STATE Get every data frame of 30 samples ($df$) \\    
    \STATE perform PCA on the new data frame ($df_{PCA}$)\\
    \STATE Project the reduced dimension data set onto the dictionary \\
    \hspace{0.5cm} $D = [D_{p}, D_{N}]$, \\
     \hspace{0.5cm} $D_{p} \in R^{l}$  and  $D_{N} \in R^{l}$,\\
      \hspace{0.5cm} l: Number of basis vectors in the dictionary,\\
    \hspace{0.5cm} $df_{PCA} = W_{p1}D_{p1}, W_{N1}D_{N1}+W_{p2}D_{p2}, W_{N2}D_{N2} +$ \\
    \hspace{0.5cm}$... +W_{pl}D_{pl}, W_{Nl}D_{Nl}$\\
    \STATE use the sparse codes ($W_{p1}, W_{N1}, ..., W_{pl}, W_{Nl} $) as input to the pre-trained classifier \\
    \STATE $S_i$ = classifier output \\
    \hspace{0.5cm}$S_i$ = 1 if poor performance is detected,\\
    \hspace{0.5cm}$S_i$ = 0 if good performance is detected)
    \ENDWHILE
  \end{algorithmic}
\end{algorithm}

\subsection{Providing Feedback for Correcting Stylistic Behavior}
\label{feedback}
\begin{figure*}[tb]
\centering
\includegraphics[width=.73\linewidth,clip ,trim={20mm 60mm 120mm 65mm}]{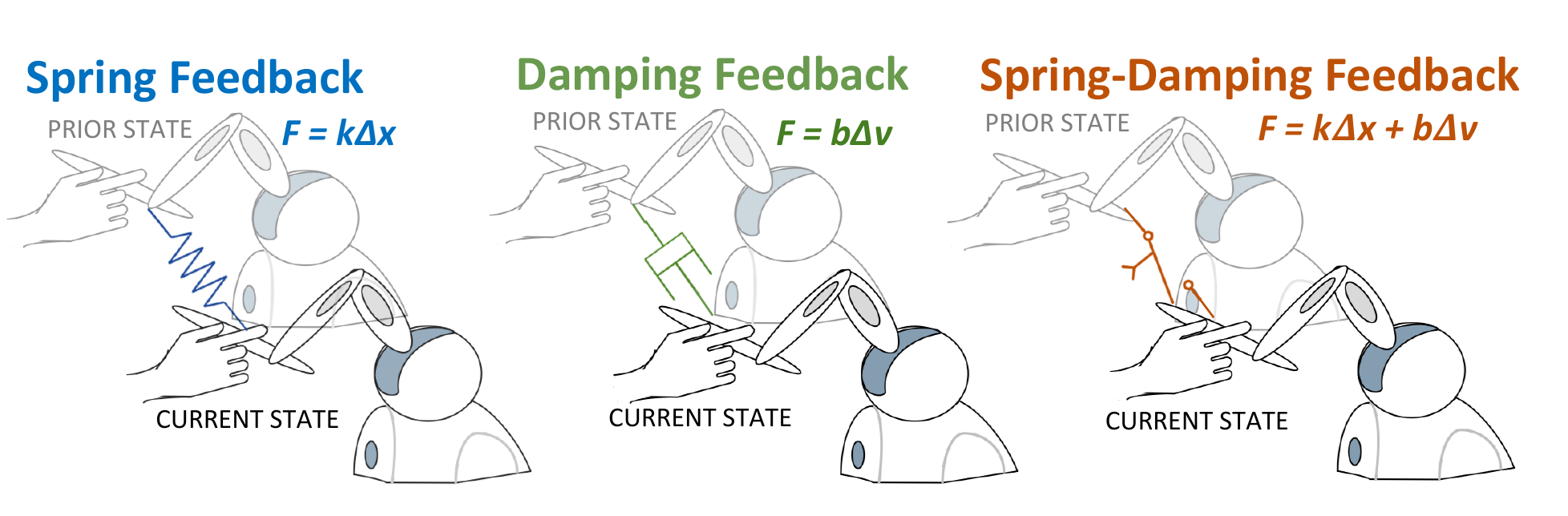} \\
\caption{Three types of haptic feedback: spring, damping, and spring + damping feedback were studied here for their ability to provide stylistic cues to the human operator. A force feedback was generated based on the user's prior position in time.}
\label{fig-feedbackl}
\vspace{-0.25cm}
\end{figure*}

To avoid confusing the operator with multiple, potentially competing feedback cause, the experiment was divided into 6 blocks and only one stylistic deficiency was detected within this set of movement trials. Based on which style detection algorithm was activated for a given block in the experiment protocol, when a poor performance was detected using the proposed near real-time algorithm, one of the three type of force feedback was turned on. 
In the following the three types of force feedback compared in this study (Fig. \ref{fig-feedbackl}) are discussed.
\begin{itemize}
\item Spring Feedback: This was calculated using the difference between the position of the hand at time t $(D_t)$, and the position at time t-1 $(D_{t-1})$. 
\begin{equation}
F_s= K_s(D_t - D_{t-1})
\end{equation}
The gain $K_s$ was obtained through trial and error and chosen to be 30. The gain was chosen to be high enough so that the user would be able to feel the feedback, but also maintain the stability of the system. This gain was fixed throughout the experiment.

\item Damping Feedback: Was calculated using the difference between the velocity of the hand at time t $(D_t)$, and the velocity at time t-1 $(D_{t-1})$. 

\begin{equation}
F_d= B_d(V_1 - V_2)
\end{equation}
The gain, $B_d$ was chosen through trial and error and was set to be 15. A lowpass filter with a cutoff frequency of 100 HZ was used to remove noise and smooth the velocity signal and prevent the system from becoming unstable.

\item Spring + Damping Feedback: Was calculated using the difference between the velocity of the hand at time t $(D_t)$, and the velocity at time t-1 $(D_{t-1})$. 

\begin{equation}
F_sd= K_{sd}(D_1 - D_2) + B_{sd}(V_1 - V_2)
\end{equation}
The gains, $K_{sd}$ and $B_{sd}$ were chosen through trial and error and set to be  10 and 5. A lowpass filter with a cutoff frequency of 100 HZ was used to remove noise and  smooth the velocity signal and prevent the system from becoming unstable.
\end{itemize}

Algorithm 2 shows the pseudo code for the haptic feedback algorithm.
\begin{algorithm}
    \caption{Feedback Generation Algorithm}
   \textbf{Input:} $S_i$: style performance , 0 otherwise  \\
   \hspace{0.5cm}$S_i = 1$ if good performance is detected,\\
   \hspace{0.5cm}$S_i = 0$ if bad performance is detected,\\
        \textbf{Output:} $f_{out}$: force feedback to be applied  \\
          \begin{algorithmic}[1]
    \STATE define feedback type (F) \\
    \hspace{0.5cm}$F = F_s$: Spring Feedback,\\
    \hspace{0.5cm}$F = F_d$: Damping Feedback,\\
    \hspace{0.5cm}$F = F_sd$: Spring+Damping Feedback\\

    \WHILE {trial not finished}
    \IF {$S_i=0$}
        \STATE $f_{out} = F$  \\
    \ENDIF
    \ENDWHILE
  \end{algorithmic}
\end{algorithm}
\section{Experimental Setup}
\label{human study}
\subsection{Data Acquisition and Simulated Task}
\begin{figure*}
\centering
\subfloat[Simulated Task]{
\includegraphics[width=0.4\linewidth,clip ,trim=320pt 180pt 320pt 175pt]{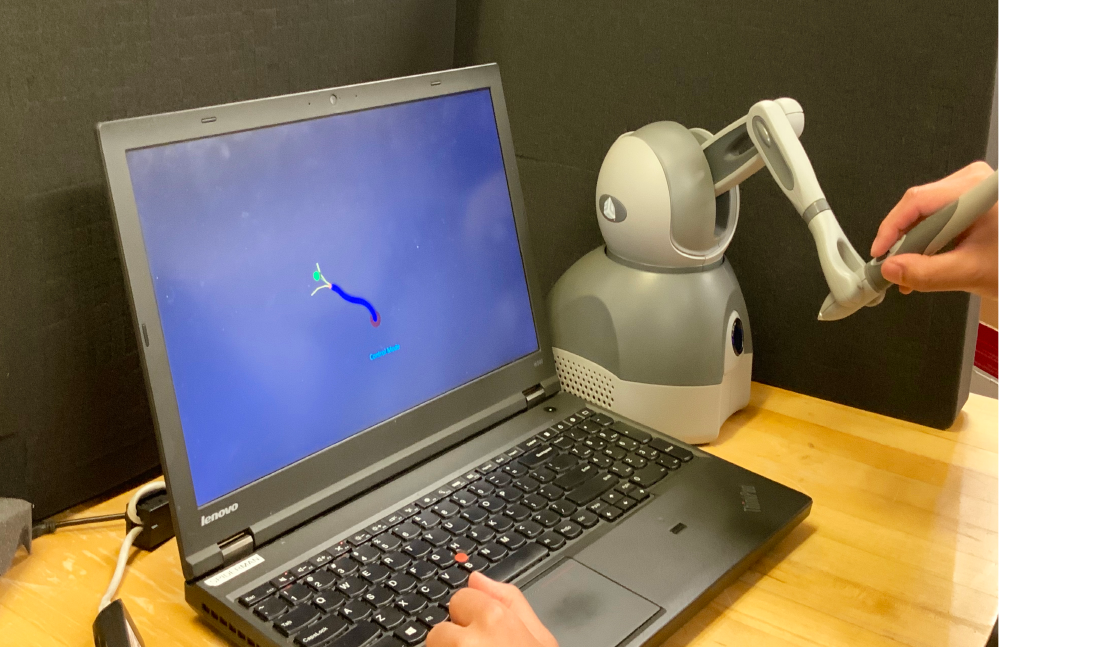}} 
\subfloat[Target Layout]{
\includegraphics[width=0.4\linewidth,clip ,trim=330pt 180pt 350pt 190pt]{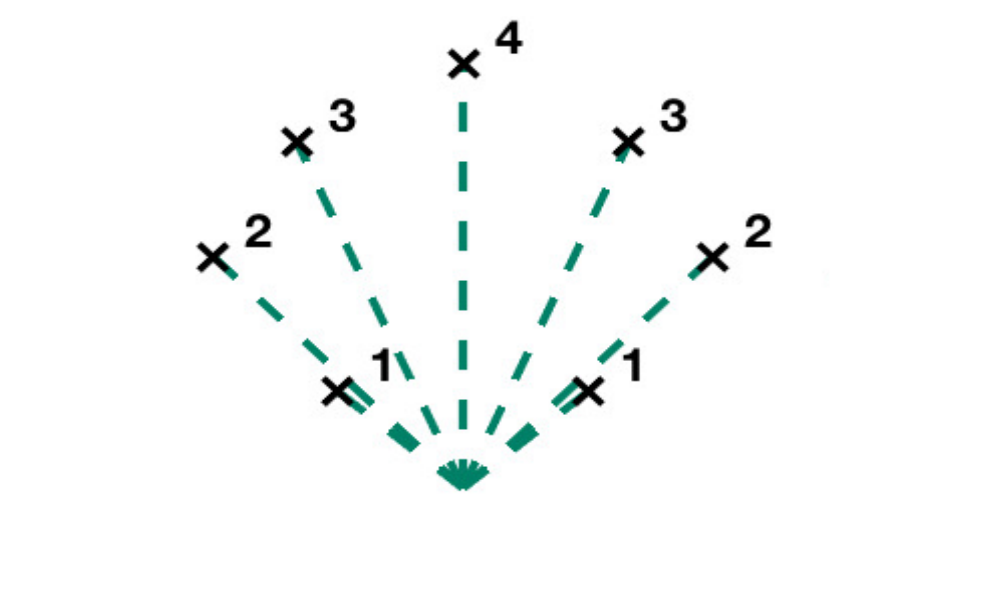}}
	\\
 \caption{(a) User interface: user interacting with simulated environment using the Geomagic Touch haptic device. The task was initiated by moving the virtual stylus to the red doughnut and would end by reaching the specified target. (b) Target layout.}
\label{fig-targettask}
\end{figure*}

The Geomagic Touch haptic device (3D Systems, Rock Hill, SC) was used in this study. This device allows for 3-degree-of-freedom force feedback and 6-degree-of-freedom sensing. It is used to both provide the user with the desired movement tasks, as well as force feedback guidance cues based on stylistic deficiencies. Position, and linear and angular velocity measurements were recorded from the stylus of the haptic device at a frequency of 256 Hz. To enable near-real-time performance, stylistic detection was performed on every frame of 30 samples of incoming data (representing 0.12 seconds).
The simulated task consisted of reaching a set of targets under a kinematically constrained environment, simulating the control of a steerable needle using Cartesian Space teleoperation~\cite{majewicz2013cartesian}. This task was chosen due to its complexity as a single-handed movement and one that naturally hunders movement is a straight-line path, which we felt would not be difficult enough to illicit stylistic changes in the user movements. The movement tasks were developed using C++ and the CHAI 3D haptic rendering library. Users were asked to reach four 5 mm targets, mirrored vertically, at predefined locations which were presented to the user at random.  The user was instructed to initialize each trial by moving the virtual stylus to the starting point. After reaching the target the user would end the trail by pressing a button on the stylus. Data was collected from the time the user initialized the haptic device until they defined the end of the trial (Fig.~\ref{fig-targettask}). 
\subsection{Experimental Protocol}
\begin{figure*}[tb]
\centering
\includegraphics[width=1\linewidth,clip ,trim={20mm 99mm 40mm 40mm}]{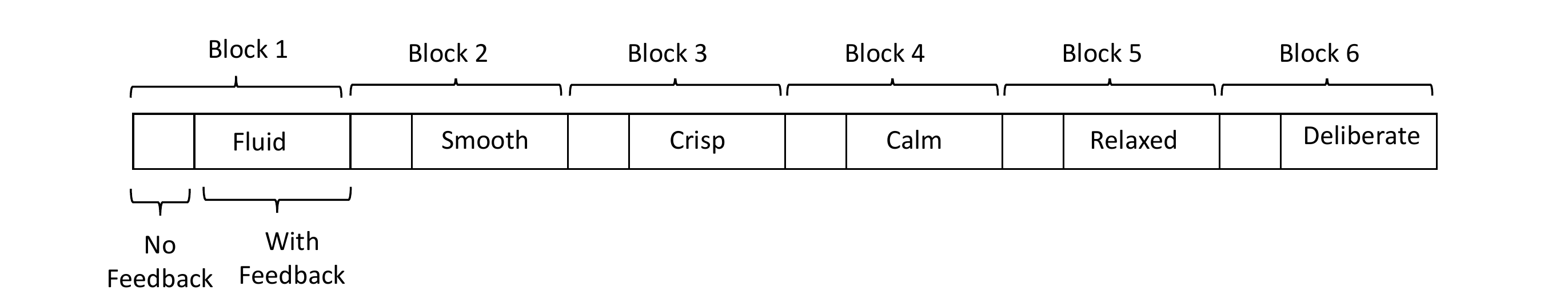} \\
\caption{An example of an experiment protocol for one subject. The protocol consists of six blocks, each related to one stylistic behavior detection algorithm that was activated for that block. For each block, the user first performed a set of reaching movements with no feedback to enable a baseline computation of style, followed by a set of trials with feedback that was provided, based on measured stylistic deficiencies. For each subject, a single feedback method was provided throughout the whole experiment, but at different points of time, depending on the style detection algorithm for that subject. Hence, a unique feedback relevant to style was provided to each subject.}
\label{fig-protocol}
\vspace{-0.25cm}
\end{figure*}

The experiment was divided into six blocks of kinematically constrained movement trials (e.g., controlling a steerable needle under cartesian space teleoperation), each block corresponding to one of the six stylistic adjectives. Each block includes a baseline segment consisting of two repetitions for each target (a total of 16 reaching trails) with no force feedback, and a segment that contains an applied force feedback for five repetitions of movements for each target (a total of 40 reaching trials). This resulted in 336 trials (6 blocks x 56 trials per block) for each participant. In each block, force feedback was provided when a stylistic deficiency was detected for the given adjective corresponding to the block. A 20 sec break was provided to the user between each block. Both target location and ordering of stylistic blocks were randomized. Figure~\ref{fig-protocol} shows an example of the experiment protocol.
\subsection{Participants}
A total of 21 subjects participated in this study. The study protocol was approved by UTD IRB office (UTD \# 14-57).  Participants had no previously reported muscular-skeletal injuries or diseases, or neurological disorders.
The subjects were divided into 3 groups of 7 subjects each. Each group was assigned the same randomized movement task, but only received either spring, damping, or spring-damping feedback for each of the stylistic adjective blocks. This parallel study design was chosen to allow us to evaluate the effect of the type of haptic feedback on corresponding changes in stylistic behavior. 

\subsection{Stylistic Behavior Performance Detection}
The kinematic data recorded from the haptic device includes user hand position, velocity, and angular velocity all in X, Y, and Z directions, resulting in 9 signal channels which are similar to the class of signals used to train the performance detection model from the JIGSAWS dataset, described in section ~\ref{Detection}. These set of  basis vectors are used here for obtaining the new spars representation of the input signal. The basis vector obtained from a class of signal similar to the input signal, better enables capturing the underlying information in the signal as opposed to using predefined dictionaries (e.g. Fourier, Wavelet, etc.)
Based on the style detection algorithm activated in each block, the new frame of data was projected onto the set of an over complete dictionary that was calculated as discussed in~\ref{dictionary}. The sparse codes for each incoming frame of data was calculated and used as input to a classifier to detect the performance quality based on the activated detection algorithm for the specific style. 
The classifier returns 0 if a poor performance is detected and returns 1 otherwise. The detection algorithms were implemented in MATLAB.
\subsection{Providing Feedback to the User}
For each frame of incoming data if a poor performance was detected (output of the classifier was 0), one type of force feedback (spring force feedback, damping force feedback, or spring-damping force feedback), was activated and applied to the user's hand.
A custom C++ code was developed to apply the force through the Geomagic Touch device. Robot Operating System (ROS) was used to build the connection between detection algorithm in MATLAB and applying the force to the user through the Geomagic Touch haptic device in C++.
Three types of forces, as discussed in section~\ref{feedback}, were studied in this experiment. Each group of subjects was provided with one type of force feedback throughout the whole experiment for all different blocks of style detection. 
\subsection{User Performance Evaluation Metrics}
To quantify the quality of performance in each trial in which a feedback was applied, the performance quantity $P$ was calculated.
For each style (i.e., each block in the protocol (Fig~\ref{fig-protocol})), the first section of the block where no feedback is applied is used as a baseline for that style. For each trial, the performance of the user was evaluated by the sum of number of times a one was detected (good performance), divided by the total number of detections in that trial. This was done for the baseline trials for each style and averaged over all force-feedback trials for the same style. 
\\
\begin{equation}
p = \dfrac{\sum_{i}^{N = 40} {(num\_positive\_WF/num\_total\_WF)/N}}{\sum_{j}^{M = 16} {(num\_positive\_NF/num\_total\_NF)/M}}
\end{equation}
\\
Where:
\textit{num\_positive\_WF} is the number of good performance detected in a trial with feedback, \textit{num\_total\_WF} is the number of total detections in a trial with feedback, \textit{i} is the trial index for feedback trials, and \textit{N} is the total number of trials with feedback for one style. In the denominator, we defined:\textit{num\_positive\_NF} as the number of good performance detected in the baseline trial (no feedback), \textit{num\_total\_NF} as the number of total detections in the baseline trial (no feedback), \textit{j} as the trial index for baseline trials, and \textit{M} as the total number of baseline trials.

\subsection{Task Performance Evaluation Metrics}
To compare the effect of the three types of feedback on the task performance, three metrics were calculated including: (1) time taken to reach the target, (2) needle trajectory straightness (the distance traveled by the needle divided by a straight line to the target), and (3) the needle position error (the distance between the needle and the target at the end of the trial).

\section{Results and Discussion}
\label{results}
We collected a total of 7056 trials (21 subjects, 336 each). Data analysis was carried out for all trials. The results include the evaluation of stylistic behavior improvement, as well as an evaluation of task performance as a function of the different types of haptic force feedback. A NASA Task Load index survey was also conducted to show how users percieved the feedback provided to them in terms of workload. 

\subsection{Effect of Force Feedback on Styles}
\begin{figure}[tb]
\centering
\includegraphics[width=1\linewidth,clip ,trim={00mm 0mm 0mm 0mm}]{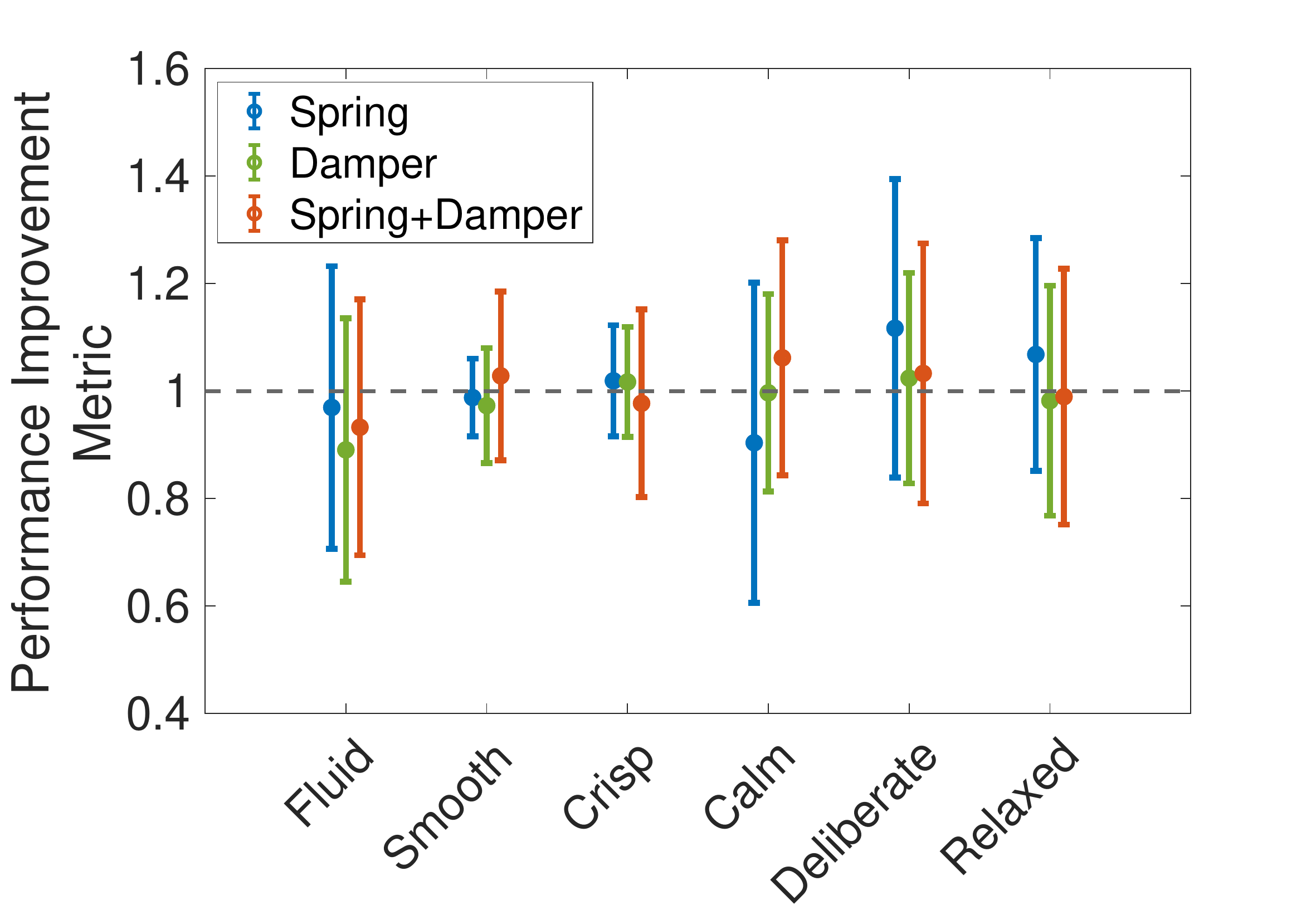} \\
\caption{Comparing the effects of three different types of haptic feedback on each style. For each group of subjects receiving the same type of feedback, the mean and standard deviation is shown for the number of positive performance normalized to the total number of detections and divided by the baseline stylistic positive performance, for each style. The values above 1 show an improvement in the performance when applying feedback compared to the no feedback condition.}
\label{Effect_feedback}
\vspace{-0.25cm}
\end{figure}

The effect of each type of force feedback on each style of movement is shown in Figure~\ref{Effect_feedback}. The mean and standard deviation of the quantity associated with good performances ($P$) for the three different types of force feedback (spring, damping, spring-damping) are plotted. This is the average number of good performances detected in the feedback segment of one block (i.e., one style detection algorithm, normalized to the average of the number of good performances detected in the baseline, no feedback, segment of the same block).
The values above the horizontal line crossing at 1 show the improvement of the movement style when applying feedback with respect to the no feedback condition and the values below this line indicate that receiving feedback did not improve the movement style compared to not receiving any feedback. 

This plot indicates that the spring force feedback was able to improve the average performance of the styles ``crisp", ``deliberate", and ``relaxed". The damping force feedback improved the ``crisp" and ``deliberate" styles on average, and the spring+damping force feedback was able to improve the ``smooth", ``calm", ``deliberate" styles on average.

Overall, all styles except for ``fluid", showed an average improvement by applying one or more types of force feedback. The ``fluid" style however showed the best performance in the absence of the forces studied here.
This can be due to the fact that other kinematic metrics, rather than the position and velocity, contribute to the fluidness of movement. In this study only force feedback associated with position and velocity were studied. According to our previous study~\cite{ershad2018}, the angular velocity of the hand movement is related to the fluidity of the movement. Thus, in future work, applying other types of force feedback which incorporate the effect of angular velocity might help to improve the fluidity of movement. This study was limited by the fact that the haptic device used was not able to provide rotational feedback cues. The style ``deliberate" was improved by all types of forces when compared to the no feedback condition; however the most improvement occurred when applying spring force feedback.
~~~~~~~~~~~~~~~~~~~~~~~~~~~~~~~~~~~~~~~~~~~~~~~~~~~~~~~~~~~~~~~~~~~~~~~~~~~~~~~

A post hoc statistical analysis was done to determine significant differences in the three types of force feedback, different targets, and task repetitions for each stylistic behavior adjective. The normality test to identify a normal distribution in the data was rejected and thus, the Kruskal Wallis test was used to identify significantly different groups. Effect significance is identified for p-values less than 0.05.

The results from the statistical analysis on different styles of movement (Table~\ref{tab-ANOVAForce}) indicate that for the styles, \textit{Fluid/Viscous, Relaxed/Tense, Deliberate/Wavering}, the \textbf{spring} force feedback showed significant difference in improving the user performance compared to the other two types of feedback.
For the \textit{Crisp/Jittery} style both the \textbf{spring} feedback and \textbf{damping} feedback showed significant improvement in performance.
For the  \textit{Calm/Anxious} styles, the \textbf{spring+damping} force feedback showed significant improvement in performance.

The statistical analysis indicate that task repetition shows no statistically significant effect on the different types of stylistic behavior as opposed to the target location which show visible statistically significant effects on different styles due to different target locations. This is shown in the third and fourth column of Table~\ref{tab-ANOVAForce}.
\begin{table*}[h]
\centering
 \small
 \caption{Statistical analysis summary of the effect of different force feedback types, targets, and task repetitions  on the stylistic behavior }
\begin{tabular}{|c|cc|cc|cc|}
	\hline
	\hline
	 \multicolumn{1}{|c|}{} &  \multicolumn{2}{c|}{Force Feedback} &  \multicolumn{2}{c|}{Target}&  \multicolumn{2}{c|}{Repetition}  \\ \cline{2-7}
	  \multicolumn{1}{|c|}{Style} & \multicolumn{1}{c}{$p$} & \multicolumn{1}{c|}{Significance} & \multicolumn{1}{c}{$p$} & \multicolumn{1}{c|}{Significance} & \multicolumn{1}{c}{$p$} & \multicolumn{1}{c|}{Significance}   \\
	\hline
	{Fluid/Viscous} 	&{ $<$0.0035}	 &  {S}$>${SD, D} &  { 0.64259}	&  {N/A} & { 0.9916} & {N/A}  \\[0.5ex]
	{Smooth/Rough}  &  { 0.2008}	&  {N/A}  &  { 0.0045}	&   {1}$>${2}   & { 0.3854} & {N/A}   \\[0.5ex]
	{Crisp/Jittery}  	 & {  $<$0.0035} & {S, D} $>$  {SD} & { 0.0045}	&   {1}$>${2,3,4} &  { 0.6953}	&  {N/A}\\[0.5ex]
	{Calm/Anxious} 	&{ $<$0.0035}	 &  {SD}$>${D}$>${S} &  { 0.2987} 	& {N/A} &   { 0.7711} 	& {N/A} \\[0.5ex]
	{Deliberate/Wavering}  &{ $<$0.0035}  & {S}$>${D, SD} & { $<$0.0035}	&  {1}$>${2}$>${3}$>${4}  & { 0.4111}	&  {N/A} \\[0.5ex]
	{Relaxed/Tense}   &{ $<$0.0035}	 &  {S}$>${D, SD}   &{ $<$0.0035} & {1}$>${2,3,4}, {2}$>${3}  & { 0.6892}	&  {N/A}   \\[0.5ex]
	\hline
	\hline

\end{tabular}
\newline
\vspace{0.1cm}
{S -- Spring, D -- Damping, SD -- Spring+Damping }
\label{tab-ANOVAForce}
\end{table*}

\subsection{Effect of Force Feedback on Task Performance}
\begin{figure*}[tb]
\centering
\includegraphics[width=1\linewidth,clip ,trim={10mm 40mm 0mm 20mm}]{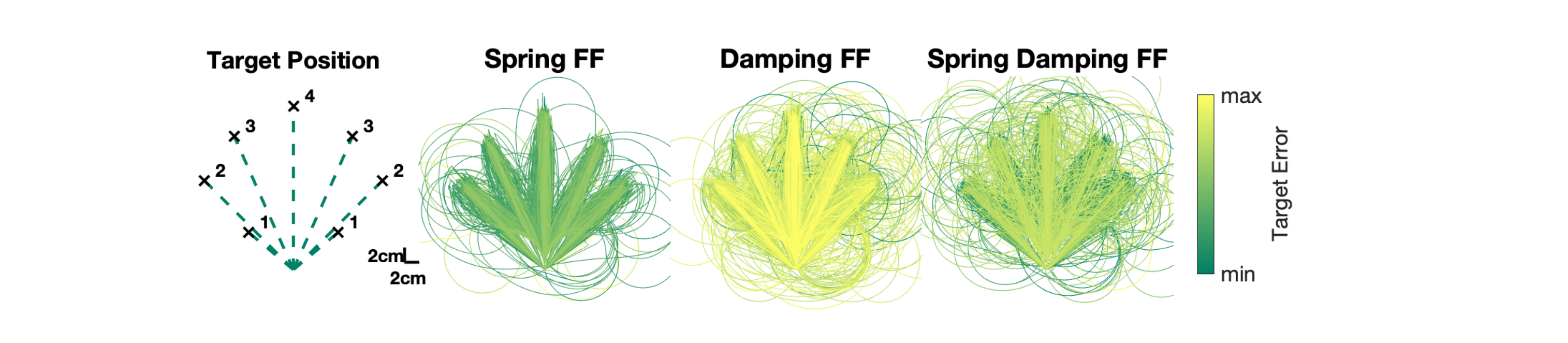} \\
\caption{Target layout and resulting needle paths for all subjects. Green paths indicating smallest error.}
\label{fig-traces}
\end{figure*}

Target configuration and needle trajectory for all trials are shown in Figure~\ref{fig-traces}, grouped by the force feedback and color-coded by target error. This figure shows the traces from all trials receiving each type of force feedback regardless of the style. The plot visually demonstrates that in general, the needle trajectory 
is more confined when applying spring force feedback compared to the other two types of feedback.

~~~~~~~~~~~~~~~~~~~~~~~~~~
	\begin{figure*}[h]
	\centering
	\subfloat[Time to Complete the Task]{
	\includegraphics[width=0.33\linewidth,clip ,trim=0pt 0pt 0pt 0pt]{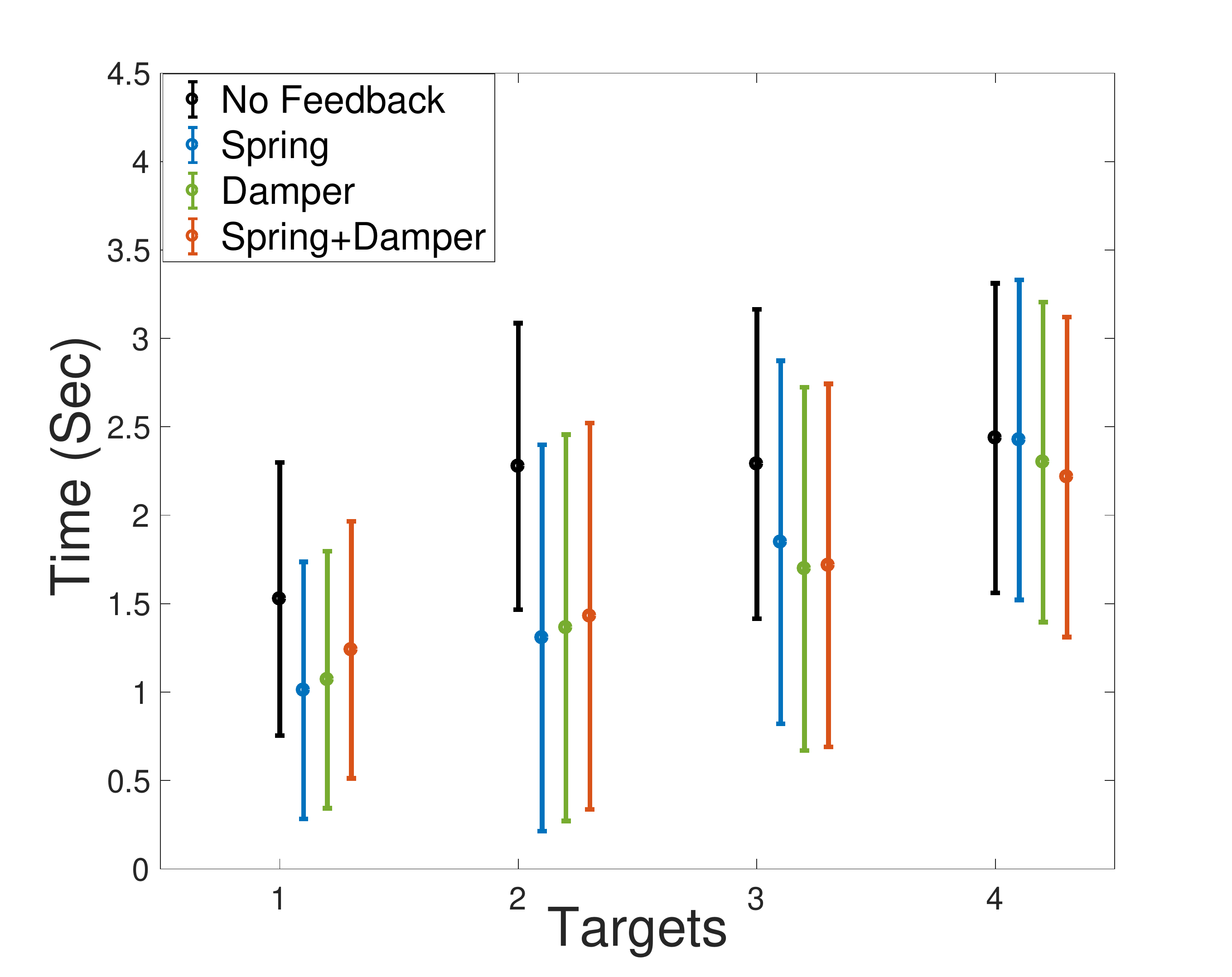}} 
	\subfloat[Target Positioning Error]{
	\includegraphics[width=0.33\linewidth,clip ,trim=0pt 0pt 0pt 0pt]{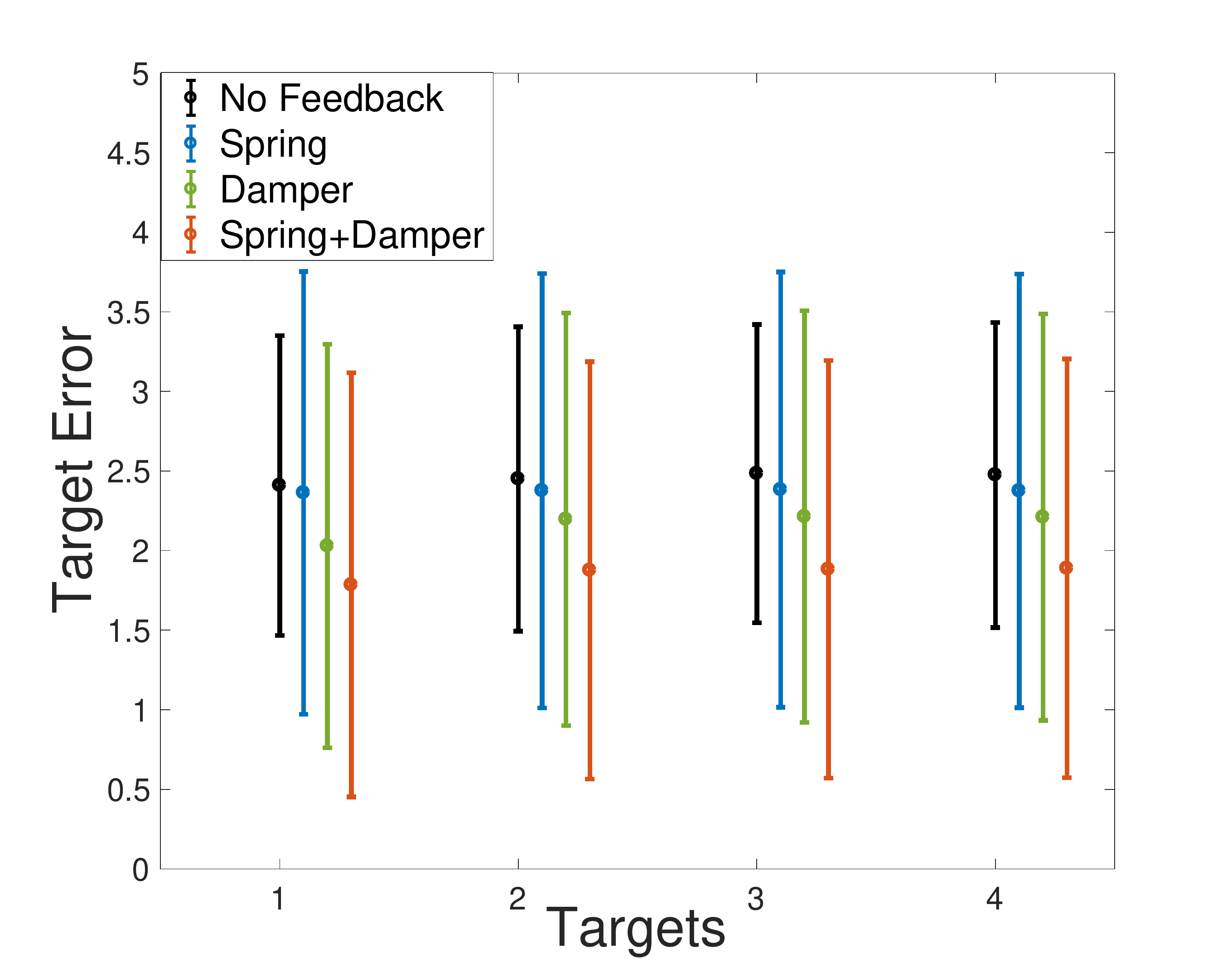}}
	\subfloat[Needle Trajectory Straightness]{
	\includegraphics[width=0.33\linewidth,clip ,trim=0pt 0pt 0pt 0pt]{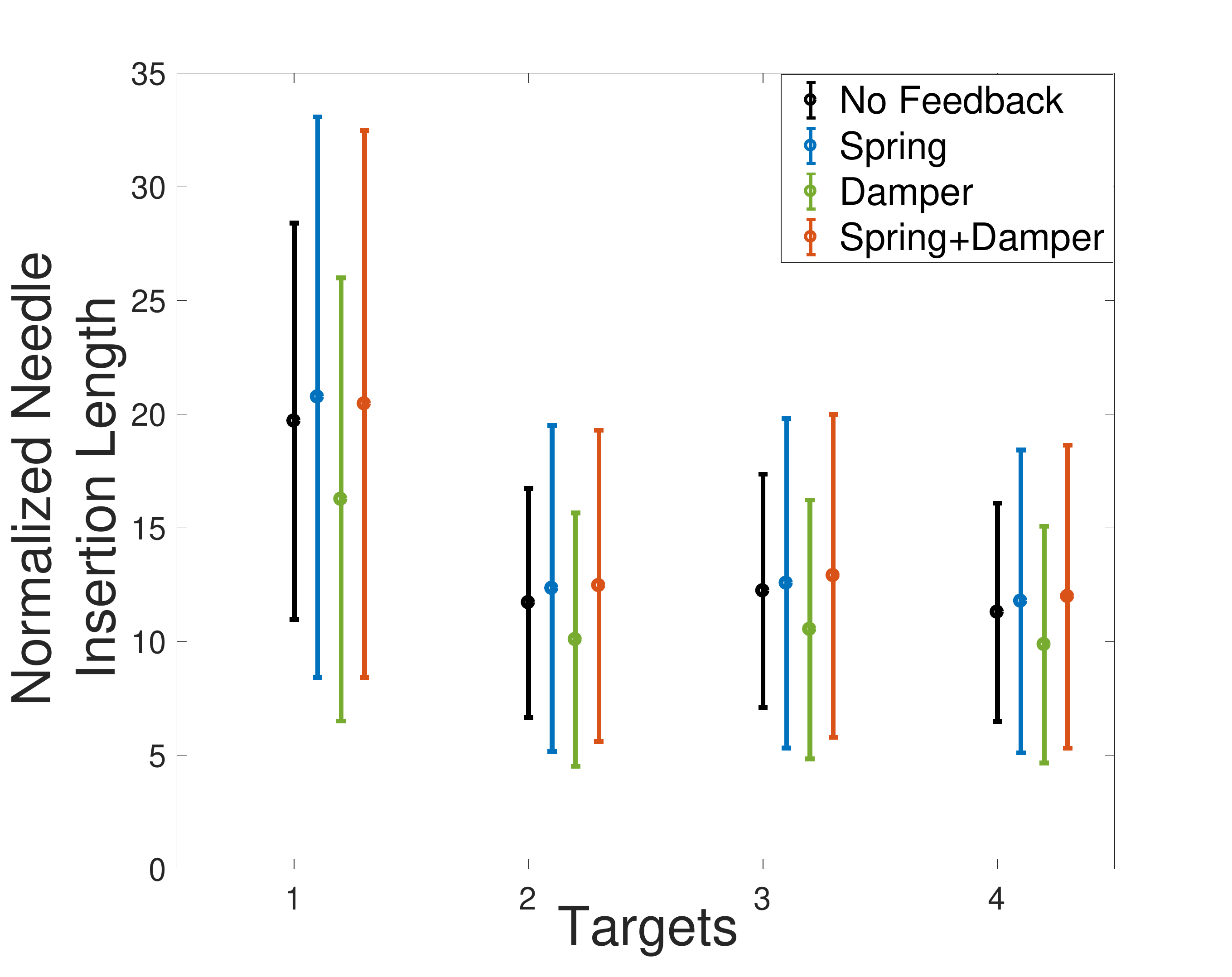}} 
	\\
 \caption{Three metrics including (a) time to complete the task, (b) target positioning error, (c) and needle trajectory straightness were used to evaluate the effect of the haptic cues on task performance. For each group (who received the same type of force feedback), the mean and standard deviation of each task performance metric is calculated and compared for 4 target locations. }
\label{fig-task performance}
\end{figure*}

For each trial, task-specific metrics including the time taken to complete the task, needle trajectory straightness, and target error, were calculated to evaluate the effects of different types of feedback, as well as the no feedback condition, on task performance. The mean and standard deviation of each of these metrics were calculated. 
Figure~\ref{fig-task performance} (a) shows that the time taken to reach the target is improved using all three types of force feedback compared to the no feedback condition. For targets 1 and 2, the group with spring force feedback showed the least time taken to reach the target, and for targets 3 and 4, the group with the spring-damping force feedback showed the least time taken to complete the task.

The target positioning error in all four targets was improved using all three groups of receiving force feedback compared to the no feedback condition; however, the group which received the spring+damping force feedback showed the least error (Fig.~\ref{fig-task performance} (b)). This indicates that applying force feedback increases the accuracy reaching task regardless of the target. 

The straightness of the trajectory traveled by the needle is compared in Fig.~\ref{fig-task performance} (c). This figure indicates that for all four targets, the group which received the damping feedback showed a straighter needle path compared to the other two feedback groups and the no feedback condition; however, spring feedback and spring+damping force feedback caused less straightness in the needle trajectory compared to the no feedback condition.

In general, the time to complete the task and the target error are improved by applying at least one of the forces. 
The straightness of the trajectory however is only improved compared to absence of feedback only when the damping feedback is applied.
A post hoc statistical analysis was also done for task performance metric, to determine significant differences in the three types of force feedback, different targets, and task repetitions. The normality test was rejected and the Kruskal Wallis test was used to identify significantly different groups. Effect significance is identified for p-values less than 0.05.
\begin{table*}
\centering
 \small
 \caption{Statistical analysis summary of the effects of force feedback types, targets, task repetition on  performance metrics}

\begin{tabular}{|c|cc|cc|cc|}
	\hline
	\hline
	 \multicolumn{1}{|c|}{} &  \multicolumn{2}{c|}{Force Feedback} & \multicolumn{2}{c|}{Target}& \multicolumn{2}{c|}{Repetition} \\ \cline{2-7}
	  \multicolumn{1}{|c|}{Performance Metric} & \multicolumn{1}{c}{$p$} & \multicolumn{1}{c|}{Significance}  & \multicolumn{1}{c}{$p$} & \multicolumn{1}{c|}{Significance} & \multicolumn{1}{c}{$p$} & \multicolumn{1}{c|}{Significance}\\
	\hline
	{Target Error} &  { 0.0389} &  { D}$>${SD} 	& { $<$0.0035} & {2}$>${3}$>${1}$>${4} & { .02033} & {N/A}  \\[0.5ex]
	{Needle Trajectory Straightness}	& {0.0398} &  {SD}$>${D}& { $<$0.0035}  & {1} $>$ {2,3,4}, {3} $>$ {4} &  {0.914}	& {N/A} \\[0.5ex]
	{Time taken to complete the task}  &  {$<$0.0035}  & {SD, D}$>${S} &  {$<$0.0035} & {2,3,4} $>$ {1} & { 0.0498} & {N/A}\\[0.5ex]
\hline
	\hline
\end{tabular}
\newline
\vspace{0.1cm}
{S -- Spring, D -- Damping, SD -- Spring+Damping }
\label{tab-ANOVA}
\end{table*}
For the task performance metrics, the statistical analysis indicates that for the target positioning error,  \textbf{spring+damping} feedback shows significant difference in reducing the error compared to \textbf{damping} feedback but is not significant compared to \textbf{spring} feedback; however, it results in a statistically significant less straighter path traveled by the needle compared to damping feedback. The \textbf{spring} feedback results in statistically significant less task completion time compared to \textbf{damping} feedback and \textbf{spring+damping} feedback (Table~\ref{tab-ANOVA}).
No statistically significant effect in task performance metrics, were found due to task repetition; however the target location shows significant importance in task-specific metrics.

\subsection{Subject Survey}
The results from user survey (NASA Task Load Index) indicate that subjects who received the spring force feedback found the feedback unpleasant and the tasks more demanding compared to subjects who received the other two types of feedback. This indicates that a haptic feedback can improve task and user performance but still be unpleasant to the user. 
\begin{figure}[h]
\centering
\includegraphics[width=1\linewidth,clip ,trim={0mm 0mm 0mm 10mm}]{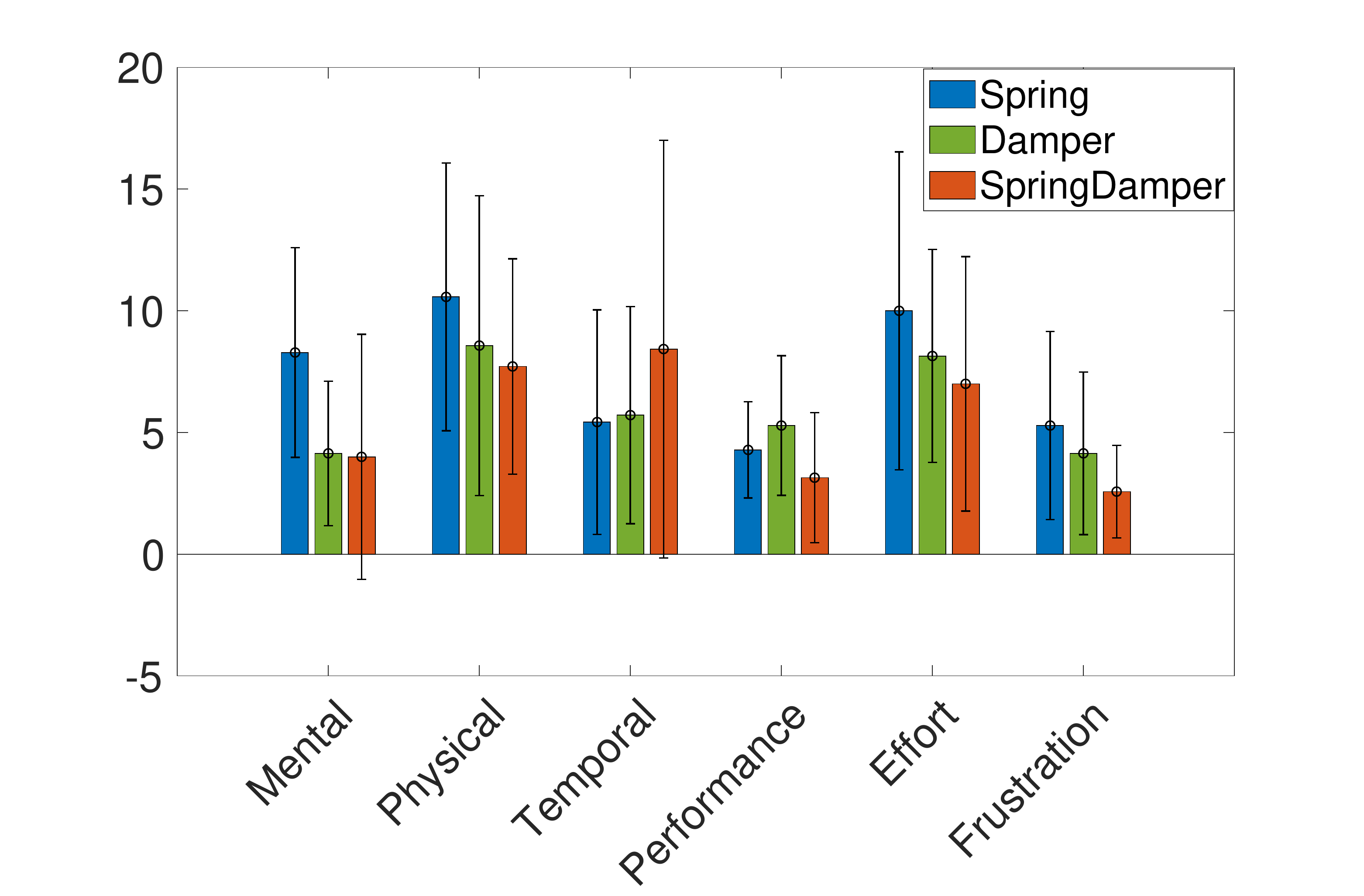} \\
\caption{NASA Task Load Index}
\label{fig-nasa}
\vspace{-0.25cm}
\end{figure}

\section{Conclusion}
\label{conclusion}
In this study we proposed an automatic training framework in a simulated environment which detects a poor behavioral performance in the user's style of movement in near-real-time and applies force feedback using a haptic device to help correct the style of movement.
We conducted a human subject study to evaluate the effect of three different types of force feedback: spring force feedback,  damping force feedback, and spring+damping force feedback on six different behavioral styles. 
The relation between the quality/style of movement and user's skill level was investigated in a previous study~\cite{ershad2018a}. 
This study builds the groundwork for using haptic as \textit{``performance feedback"} (as opposed to reflective or guidance feedback) for improving stylistic behavior and hence, quality of movement.

The results indicated that ``Spring" force feedback resulted in less time to complete the task, hence faster performance speed.
It also helped demonstrate a more fluid, crisp, calm, and deliberate behavior in the user's movement.

``Spring+Damping'' feedback reduced the target error resulting in a more accurate performance; however, it resulted in a less straight needle path and more time to complete the task. It helped demonstrate a more relaxed performance.

``Damping" force feedback resulted in a straighter line traveled by the needle in the simulated task towards the target; however, it lead to an increased target error and a slower speed resulting in more time taken to complete the task. It helped the user to demonstrate a more crisp performance.

In this study we considered only three type of force feedback related to position and velocity; however, this might not be sufficient for all styles, since other kinematic metrics can also be associated with some styles. In future studies, considering other types of force feedback can help improve the performance of the styles that were not improved using only a position or velocity force feedback.

Another area of focus in the future will be to evaluate the effectiveness of the proposed haptic training method for improving stylistic behavior in long term. For this purpose, subjects will be trained in three groups, each group will be receiving a different type of haptic feedback and the their performance will be monitored and measured over time.

Furthermore this study was carried out in a simulation environment which has limitations in representing a real surgical task. Future studies will focus on addressing this issue by implementing the proposed method on the da Vinici research kit (dvrk) along with a training task.

This study provides the groundwork for continued research on user performance based feedback for adaptive training.

\appendices

\section*{Acknowledgment}
The authors would like to thank Ziheng Wang for his contributions develop a bridge between MATLAB and ROS. 
 
The research reported in this publication was supported by the National Center for Advancing Translational Sciences of the National Institutes of Health under award Number UL1TR001105. The content is solely the responsibility of the authors and does not necessarily represent the official views of the NIH.

%

%
\bibliographystyle{IEEEtran.bst}
\bibliography{references}
\end{document}